\def\year{2017}\relax
\documentclass[letterpaper]{article}
\usepackage{aaai17}
\usepackage{times}
\usepackage{helvet}
\usepackage{courier}

\usepackage{amsthm}

\usepackage{amsmath, amssymb}
\usepackage{algorithm,algorithmic} 
\newtheorem{thm}{Theorem}
\newtheorem{prop}{Proposition}
\newtheorem{lemma}{Lemma}

\usepackage{graphicx}
\usepackage{subfigure}

\def \A {\mathcal{A}}
\def \X {\mathcal{X}}

\def \R {\mathbb{R}}

\def \w {\mathbf{w}}
\def \v {\mathbf{v}}
\def \x {\mathbf{x}}

\def \E {\mathrm{E}}

\def \x {\mathbf{x}}

\def \diag {\mbox{diag}}

\def \y {\mathbf{y}}

\def \P {\mathcal{P}}
\def \Uh {\widehat{U}}

\def \xh {\widehat{\x}}
\def \wh {\widehat{\w}}
\def \vh {\widehat{\v}}

\def \ah {\widehat{\alpha}}

\def \Xh {\widehat{X}}

\begin{document}
\title{Efficient Non-oblivious Randomized Reduction for Risk Minimization with Improved Excess Risk Guarantee}
\author{Yi Xu$^1$, Haiqin Yang$^2$, Lijun Zhang$^3$, Tianbao Yang$^1$\\
$^1$Department of Computer Science, The University of Iowa, Iowa City, IA 52242, USA\\
$^2$Department of Computing, Hang Seng Management College, Hong Kong\\
$^3$National Key Laboratory for Novel Software Technology, Nanjing University, Nanjing 210023, China\\
\{yi-xu, tianbao-yang\}@uiowa.edu, hqyang@ieee.org, zhanglj@lamda.nju.edu.cn\\
}
\maketitle
\begin{abstract}
In this paper, we address learning problems for high dimensional data.  Previously, oblivious random projection based approaches that project high dimensional features onto a random subspace have been used in practice for tackling high-dimensionality challenge in machine learning.  Recently, various non-oblivious randomized reduction methods have been developed and deployed for solving many numerical problems such as matrix product approximation, low-rank matrix approximation, etc.  However, they are less explored for the machine learning tasks, e.g., classification.  More seriously, the theoretical analysis of excess risk bounds for risk minimization, an important measure of generalization performance, has not been established for non-oblivious randomized reduction methods.  It therefore remains an open problem what is the benefit of using them over previous oblivious random projection based approaches.  To tackle these challenges, we propose an algorithmic framework for employing non-oblivious randomized reduction method for general empirical risk minimizing in machine learning tasks, where the original high-dimensional features are projected onto a random subspace that is derived from the data with a small matrix approximation error.  We then derive the first excess risk bound for the proposed non-oblivious randomized reduction approach without requiring strong assumptions on the training data. The established excess risk bound exhibits that the proposed approach provides much better generalization performance and it also sheds more insights about different randomized reduction approaches.  Finally, we conduct extensive experiments on both synthetic and real-world benchmark datasets, whose dimension scales to $O(10^7)$, to demonstrate the efficacy of our proposed approach.
\end{abstract}

\section{Introduction}
Recently, the scale and dimensionality of data associated with machine learning and data mining applications have seen unprecedented growth, spurring the BIG DATA research and development.  Learning from large-scale ultrahigh-dimensional data remains a computationally challenging problem.  The big size of data not only increases the memory footprint but also increases the computational costs pertaining to optimization.  A popular approach for addressing the high-dimensionality challenge is to perform dimensionality reduction. Nowadays, randomized reduction methods are emerging to be attractive for dimensionality reduction.  Compared with traditional dimensionality reduction methods (e.g., PCA and LDA), randomized reduction methods  (i)  can lead to simpler algorithms that are easier to analyze~\cite{mahoney2011randomized}; (ii) can often be organized to exploit modern computational architectures better than classical dimensional reduction methods~\cite{halko2011finding}; (iii) can be more efficient without loss in efficacy~\cite{paul2013random}.

Generally, randomized reduction methods can be cast into two types: the first type of methods reduces a set of high-dimensional vectors into a low dimensional space independent of each other. These methods usually sample a random matrix independent of the data and then use it to reduce the dimensionality of the data.  The second type of methods projects a set of vectors (in the form of a matrix) onto a subspace such that the original matrix can be well reconstructed from the projected matrix and the subspace.  Therefore, the subspace to which the data is projected depends on the original data.  These methods have been deployed for solving many numerical problems related to matrices, e.g., matrix product approximation, low-rank matrix approximation, approximate singular value decomposition~\cite{DBLP:journals/siammax/BoutsidisG13,halko2011finding}. To differentiate these two types of randomized reduction methods, we refer to the first type as oblivious randomized reduction, and refer to the second type as non-oblivious randomized reduction. We note that in literature oblivious and non-oblivious are used interchangeably with  data-independent and data-dependent. Here, we use the terminology commonly appearing in matrix analysis and numerical linear algebra due to that the general excess risk bound depends on the matrix approximation error.

However, we have not seen any comprehensive study on the statistical property (in particular the excess risk bound) of these randomized reduction methods applied to risk minimization in machine learning.  The excess risk bound measures the generalization performance of a learned model compared to the optimal model from a class that has the best generalization performance.  The excess risk bounds facilitate a better understanding of different learning algorithms and have the potential to guide us to design better algorithms~\cite{DBLP:conf/icml/KuklianskyS15}.  It is worth noting that several studies have been devoted to understanding the theoretical properties of oblivious randomized reduction methods applied to classification and regression problems. For example, \cite{blum2005random,shi2012margin,paul2013random} analyzed the  preservation of the margin of SVM based classification methods with  randomized dimension reduction. \cite{DBLP:journals/tit/0005MJYZ14,yang2015theory,DBLP:journals/tit/PilanciW15} studied the problem from the perspective of optimization. Nonetheless,  these results are limited in the sense that (i) they focus on only oblivious randomized reduction where the data is projected onto a random subspace independent of the data; (ii) they depend heavily on strong assumptions of the training data or the problem, e.g., low-rank of the data matrix, linear separability of training examples, or the sparsity of optimal solution, and (iii) some of these results do not directly carry over to  the excess risk bounds.

To tackle the above challenges, we propose an algorithmic framework for employing non-oblivious randomized reduction (NOR) method to project the original high-dimensional features onto a random subspace that is derived from the original data.  We study and establish the excess risk bound of the presented randomized algorithms for risk minimization.  Different from previous results for oblivious randomized reduction methods, our theoretical analysis does not require assumptions of the training data or the problem, such as low-rank of the data matrix, linear separability of training examples, and the sparsity of optimal solution.  When the data matrix is of low-rank or has a fast spectral decay, the excess risk bound of NOR is much better than that of oblivious randomized reduction based methods. Empirical studies on synthetic and real data sets corroborate  the theoretical results and demonstrate the effectiveness of the proposed methods.

\section{Related Work} \label{ralatedwork}
In literature, tremendous studies are devoted to non-oblivious randomized reduction in matrix applications. The focus of these studies is to establish matrix approximation error  or the recovery error of the solution (e.g., in least-squares regression). Few studies have examined their properties for risk minimization in machine learning. For oblivious randomized reduction methods, there exist some theoretical work trying to understand their impact on prediction performance~\cite{blum2005random,shi2012margin,paul2013random}. This work differentiates from these studies in that we focus on the statistical property (the generalization property) of non-oblivious randomized reduction for expected risk minimization.

We employ tools in statistical learning theory to study the excess risk bounds of randomized reduction and results from randomized matrix theory to understand the order of the excess risk bound. A popular method for expected risk minimization is regularized empirical risk minimization~\cite{citeulike:106699}. The excess risk bounds of regularized empirical risk minimization have been well understood. In general, given a sample of size $n$ it can achieve a risk bound of $O(1/\sqrt{n})$. Under some special conditions (e.g., low noise condition) this bound can be further improved~\cite{BousquetBL03}. However, it is still  not entirely clear what is the order of excess risk for learning from randomized dimensionality reduced data. The recovery result from~\cite{DBLP:journals/tit/0005MJYZ14,yang2015theory} could end up with  an order of $O(1/\sqrt{m})$ excess risk for oblivious randomized reduction, where $m$ is the reduced dimensionality. However, it relies on strong assumptions of the data. \cite{DBLP:conf/icml/DurrantK13} proved  the generalization error of the linear classifier trained on randomly projected data by oblivious randomized reduction, which is upper bounded by the training error of the classifier learned in the original feature space by empirical risk minimization plus the VC-complexity in the projection space (proportional to $O(1/\sqrt{m})$ and plus terms depending on the average flipping probabilities on the training points defined as $(1/n)\sum_{i=1}^n\Pr(sign(\w_n^TA^{\intercal}A\x_i)\neq sign(\w_n^{\intercal}\x_i))$, where $\w_n$ is a model learned from the original data by empirical risk minimization. However, the order of the average flipping probabilities is generally unknown.

Random sampling  (in particular uniform sampling) has been used in the Nystr\"{o}m method for approximating a big kernel matrix. There are some related work focusing on the statistical properties of the Nystr\"{o}m based kernel method~\cite{DBLP:conf/nips/YangLMJZ12,Bach2012,DBLP:journals/tit/JinYMLZ13,alaoui2015fast}. We note that the presented empirical risk minimization with non-oblivious randomized reduction using random sampling is similar to using the Nystr\"{o}m  approximation on the linear kernel. However,  in the present work  besides random sampling, we also study other efficient  randomized reduction methods using different random matrices. By leveraging recent results of these randomized reduction methods we are able to obtain better performance than using random sampling.

\section{Preliminaries}\label{sec:pre}
Let $(\x, y)$ denote a feature vector and a label that follow a distribution $\P=\P(\x, y)$, where $\x\in\X\subset\R^d$ and $y\in\mathcal Y$. In the sequel, we will focus on $\mathcal Y=\{+1, -1\}$ and $\mathcal Y=\R$. However, we emphasize that the results are applicable to other problems (e.g., multi-class and multi-label classification).  We denote by $\ell(z, y)$ a non-negative loss function that measures the inconsistency between a prediction $z$ and the label $y$.  Let $\w \in \R^d$, then by assuming a linear model  $z = \w^{\intercal}\x$ for prediction, the risk minimization problem in machine learning is to solve following problem:
\begin{align}
\w_* = \arg\min_{\w\in\R^d}\E_{\P}[\ell(\w^{\intercal}\x, y)]
\end{align}
where $\E_{\P}[\cdot]$ denotes the expectation over $(\x, y)\sim \P$.

Let $\A$ be an algorithm that learns an approximate solution $\w_n$ from a sample  of size $n$, i.e., $\{(\x_1,y_1), \ldots, (\x_n, y_n)\}$. The excess risk of $\w_n$ is defined as the difference between the expected risk of the solution $\w_n$ and that of the optimal solution $\w_*$:
\begin{align}
\text{ER}&(\w_n, \w_*) = \E_{\P}[\ell(\w_n^{\intercal}\x, y)] -  \E_{\P}[\ell(\w_*^{\intercal}\x, y)]
\end{align}
A popular method for learning an approximate solution $\w_n$  is based on regularized empirical risk minimization (ERM), i.e.,
\begin{align}\label{eqn:org}
\w_n =\arg\min_{\w\in\R^d} \frac{1}{n}\sum_{i=1}^n\ell(\w^{\intercal}\x_i, y_i) + \frac{\lambda}{2}\|\w\|_2^2
\end{align}
The ERM problem is sometimes solved by solving its dual problem:
\begin{align}\label{eqn:orgd1}
\alpha_* =\arg\max_{\alpha\in\R^n} -\frac{1}{n}\sum_{i=1}^n\ell_i^*(\alpha_i) - \frac{1}{2\lambda n^2}\alpha^{\intercal}X^{\intercal}X\alpha
\end{align}
where $\alpha$ is usually called dual variable, $\ell_i^*(\alpha)=\max_{z}\alpha z - \ell(z, y_i)$ is the conjugate dual of the loss function, and $X=(\x_1,\ldots, \x_n)\in\R^{d\times n}$ is the data matrix. With $\alpha_*$, we have $\w_n = - \frac{1}{\lambda n}X\alpha_*$.

\section{Oblivious Randomized Reduction}\label{sec:ob}
In this section, we present an excess risk bound of oblivious randomized reduction building  on previous theoretical results to facilitate the comparison with our result of non-oblivious randomized reduction. The idea of oblivious randomized reduction is to reduce a high-dimensional feature vector $\x\in\R^d$ to a low dimensional vector by $\xh=A\x\in\R^{m}$, where $A\in\R^{m\times d}$ is a random matrix that is independent of the data. A traditional approach is to use a Gaussian matrix with each entry independently sampled from a normal distribution with mean zero and variance $1/m$~\cite{dasgupta-2003-jl}. Recently, many other types of random matrix $A$ are proposed that lead to much more efficient computation of reduction, including subsampled randomized {Hadamard transform} (SRHT)~\cite{DBLP:journals/siammax/BoutsidisG13} and random hashing (RH)~\cite{Kane:2014:SJT:2578041.2559902}. The key property of $A$ that plays an important role in the analysis is  that it should preserve the Euclidean length of a high-dimensional vector with a high probability, which is stated formally in Johnson-Lindenstrauss (JL) lemma below.
\begin{lemma}[JL Lemma]\label{lemma:1}
For  any $0 < \epsilon$, $\delta < 1/2$, there exists a probability distribution on matrices $A\in\R^{m\times d}$ such that there exists a small universal constant $c>0$ and for any fixed $\x\in\R^d$, with a probability at least $1-\delta$, we have
\begin{align*}
\left|\|A\x\|_2^2 - \|\x\|^2_2\right|\leq c\sqrt{\frac{\log(1/\delta)}{m}}\|\x\|_2^2
\end{align*}
\end{lemma}
The key consequence of the JL lemma is that we can reduce a set of $d$-dimensional vectors into a low dimensional space with a reduced dimensionality independent of $d$ such that the pairwise distance between any two points can be well preserved.

Given the JL transform $A\in\R^{m\times d}$, the problem can be imposed as,
\begin{align}\label{eqn:r}
\min_{\v\in\R^m}\E_{\P}[\ell(\v^{\intercal}A\x, y)]
\end{align}
Previous studies have focused on using the ERM of the above problem
\begin{align}
\min_{\v\in\R^m}\frac{1}{n}\sum_{i=1}^n\ell(\v^{\intercal}\xh_i, y_i) + \frac{\lambda}{2}\|\v\|_2^2
\end{align}
to learn a model in the reduced feature space or using its dual solution to recover a model in the original high-dimensional space:
\begin{align}\label{eqn:orgd}
\ah =\arg\max_{\alpha\in\R^n} -\frac{1}{n}\sum_{i=1}^n\ell_i^*(\alpha_i) -  \frac{1}{2\lambda n^2}\alpha^{\intercal}\Xh^{\intercal}\Xh\alpha
\end{align}
where $\Xh=(\xh_1,\ldots, \xh_n)\in\R^{m\times n}$. For example, \cite{DBLP:journals/tit/0005MJYZ14} proposed a dual recovery approach to recover a model in the original high-dimensional space that is close to the optimal solution $\w_n$ in~(\ref{eqn:org}). The dual recovery approach consists of two steps (i) the first step obtains  an approximate  dual solution $\ah\in\R^n$ by solving the dual problem in~(\ref{eqn:orgd}),  and (ii) the second step  recovers  a high-dimensional model  by $\wh_n = - \frac{1}{\lambda n}X\ah$. By making a low-rank assumption of the data matrix, they established a recovery error $\|\w_n - \wh_n\|_2$ in the order of $O(\sqrt{r/m}\|\w_n\|_2)$, where $r$ represents the rank of the data matrix. The theory has been generalized to full rank data matrix but with an additional assumption that the optimal primal solution $\w_n$ or the optimal dual solution $\alpha_*$ is sparse~\cite{DBLP:journals/tit/0005MJYZ14,yang2015theory}. A similar order  $O(\sqrt{r/m}\|\w_n\|_2)$ of recovery error was established, where $r$ represents the number of non-zero elements in the optimal solution.  The proposition below exhibits the excess risk bound building on the recovery error.
\begin{prop}\label{prop:1}
Suppose $\ell(z, y)$ is Lipschitz continuous and $A$ is a JL transform.  Let $\wh_n$ denote a recovered model by an ERM approach such that $\|\w_n - \wh_n\|_2\leq O(\sqrt{r/m}\|\w_n\|_2)$. Then
\begin{align*}
\text{ER}(\wh_n, \w_*)&\triangleq\E_{\P}[\ell(\wh_n^{\intercal}\x, y)] -  \E_{\P}[\ell(\w_*^{\intercal}\x, y)]\\
&\leq O(\sqrt{r/m}\|\w_n\|_2 + 1/\sqrt{n})
\end{align*}
\end{prop}
{\bf Remark:} In the above bound, we omit dependence on upper bound of the data norm $\|\x\|\leq R$.
Although the synthesis of  the proposition and previous recovery error analysis can give us guarantee on the excess risk bound, it relies on certain assumptions of the data, which may not hold in practice.

\section{Non-Oblivious Randomized  Reduction} \label{methodRANOR}
The key idea of non-oblivious randomized reduction is to compute a subspace $\widehat U\in\R^{d\times m}$ from the data matrix $X\in\R^{d\times n}$ such that the projection of the data matrix to the subspace is close to the data matrix. To compute the subspace $\Uh$, we first sample a random matrix $\Omega\in\R^{n\times m}$ and compute $Y=X\Omega\in\R^{d\times m}$. Then let $\widehat U$ be the left singular vector matrix of $Y$. This technique has been used in low-rank matrix approximation, matrix product approximation and approximate singular value decomposition (SVD) of a large matrix~\cite{halko2011finding}. Various random matrices $\Omega$ can be used as long as the matrix approximation error defined below can be well bounded.
\begin{align}\label{eqn:approx}
\|X- \widehat U\widehat U^{\intercal}X\|_2=\|X - P_YX\|_2
\end{align}
where $P_Y=\Uh\Uh^{\intercal}$ denotes the projection to the subspace $\Uh$.  We defer more discussions on different random matrices and their impact on the excess risk bound to subsection ``Matrix Approximation Error".

Next, we focus on the non-oblivious reduction defined by $\widehat U$ for risk minimization. Let  $\xh_i = \widehat U^{\intercal}\x_i$ denote the reduced feature vector and  $\Xh = \Uh^{\intercal}X\in\R^{m\times n}$ be the reduced data matrix. We propose to solve the following ERM problem:
\begin{align}\label{eqn:redprob-2}
\widehat\v_n=\arg\min_{\v\in\R^m}\frac{1}{n}\sum_{i=1}^n\ell(\v^{\intercal}\xh_i, y_i) +  \frac{\lambda}{2}\|\v\|_2^2.
\end{align}
\begin{algorithm}[tb]
\small
   \caption{ERM with Non-Oblivious  Randomized  Reduction (NOR)}
   \label{alg:RDDR}
\begin{algorithmic}[1]
   \STATE Compute $Y = X\Omega\in\R^{d\times m}$, where $\Omega\in\R^{n\times m}$ is a random subspace embedding matrix
   \STATE Compute SVD of $Y=\widehat U\widehat\Sigma \widehat V^{\intercal}$, where $\widehat U\in\R^{d\times m}$
   \STATE Compute the reduced data by $\Xh = \widehat U^{\intercal}X\in\R^{m\times n}$
   \STATE Solve the reduced problem in Eqn.~(\ref{eqn:redprob-2})
   \STATE Output $\vh_n$
\end{algorithmic}
\end{algorithm}
To understand the non-oblivious randomized reduction for ERM, we first see that the problem above is equivalent to
\begin{align*}
\min_{\w=\widehat U\v, \v\in\R^m}\frac{1}{n}\sum_{i=1}^n\ell(\w^{\intercal}\x_i, y_i) +  \frac{\lambda}{2}\|\w\|_2^2.
\end{align*}
due to that $\|\Uh\v\|_2=\|\v\|_2$. Compared to~(\ref{eqn:org}), we can see that the ERM with non-oblivious randomized reduction is restricting the model $\w$ to be $\widehat U\v$. Since $\widehat U$ can capture the top column space of $X$ due to the way it is constructed, and therefore the resulting model $\wh_n = \widehat U\vh_n$ is close to the top column space of $X$. Thus, we expect $\wh_n$ to be  close to $\w_n$.  The procedure is described in details in Algorithm~\ref{alg:RDDR}. We note that the SVD in step 2 can be computed efficiently for sparse data. We defer the details into the appendix.

\subsection{Excess Risk Bound}
Here, we show an excess risk bound of the proposed ERM with non-oblivious randomized reduction.  The logic of the analysis is to first derive the optimization error of the approximate model $\wh_n = \widehat U\vh_n$ and then explore the statistical learning theory to bound the excess risk. In particular, we will show that the optimization error and consequentially the excess risk is bounded by the matrix approximation error in~(\ref{eqn:approx}). To simplify the presentation, we introduce some notations:
\begin{align}\label{eqn:l2}
F(\w) &=\frac{1}{n}\sum_{i=1}^n\ell(\w^{\intercal}\x_i, y_i) + \frac{\lambda}{2} \|\w\|^2\\
\bar F(\w) &= \E_{\P}[\ell(\w^{\intercal}\x, y)] + \frac{\lambda}{2}\|\w\|_2^2
\end{align}
Next, we derive the optimization error of $\wh_n = \Uh\vh_n$.
\begin{lemma}\label{lem:1}
Suppose the loss function is $G$-Lipschitz continuous.  Let  $\wh_n = \widehat  U\vh_n$.
 We have
\[
F(\wh_n)\leq F(\w_n) + \frac{G^2}{2\lambda n}\|X- P_YX\|^2_2
\]
\end{lemma}
The lemma below bounds the excess risk by  the optimization error.
\begin{lemma}[Theorem 1~\cite{DBLP:conf/nips/SridharanSS08}]
Assume the loss function is $G$-Lipschitz continuous and $\|\x\|_2\leq R$. Then, for any $\delta>0$ and any $a > 0$, with probability at least $1-\delta$, we have that for any $\w_*\in\R^d$
\begin{align*}
\bar F(\wh_n) - \bar F(\w_*)&\leq (1+a) (F(\wh_n) - F(\w_n)) \\
& + \frac{8(1+1/a)G^2R^2(32 + \log(1/\delta))}{\lambda n}
\end{align*}
\end{lemma}
Using the lemma above and the the result in Lemma~\ref{lem:1}, we have the following theorem about the excess risk bound.
\begin{thm}
Assume the loss function is $G$-Lipschitz continuous and $\|\x\|_2\leq R$. Then, for any $\delta>0$ and any $a > 0$, with probability at least $1-\delta$, we have that for any $\w_*$ such that $\|\w_*\|_2\leq B$
\begin{align*}
\text{ER}(\wh_n, \w_*)&\leq \frac{\lambda B^2}{2}  +\frac{G^2(1+a)}{2\lambda n}\|X- P_YX\|^2_2\\
&+  \frac{8(1+1/a)G^2R^2(32 + \log(1/\delta))}{\lambda n}
\end{align*}
In particular, if we optimize $\lambda$ over the R.H.S., we obtain
\begin{align*}
\text{ER}(\wh_n, \w_*)&\leq   \frac{GB\sqrt{(1+a)}}{\sqrt{n}}\|X- P_YX\|_2\\
&+  \frac{4GRB\sqrt{(1+1/a)(32 + \log(1/\delta))}}{\sqrt{n}}
\end{align*}
\end{thm}
{\bf Remark:} Note that the above theorem bounds the excess risk by the matrix approximation error. Thus, we can leverage state-of-the-art results on the matrix approximation to study the excess risk bound. Importantly, future results about matrix approximation can be  directly plugged into the excess risk bound. When the data matrix is of low rank $r$, then if $m\geq \Omega(r\log r)$ the matrix approximation error can be made zero (see below). As a result, the excess risk bound of $\wh_n$ is $O(1/\sqrt{n})$, the same to that of $\w_n$. In contrast, the excess risk bound in Proposition~\ref{prop:1} of oblivious randomized reduction for ERM is $O(\sqrt{r/m})$ for the dual recovery approach under the low rank assumption.

\subsection{Matrix Approximation Error}\label{sec:mae}
In this subsection, we will present some recent results on the matrix approximation error of four commonly used randomized reduction operators $\Omega\in\R^{n \times m}$, i.e., random sampling (RS), random Gaussian  (RG), subsampled  randomized Hadamard transform (SRHT), and random hashing (RH), and discuss  their impact on the excess risk bound. More details of these four randomized reduction operators can be found in~\cite{yang2015theory}. We first introduce some notations used in matrix approximation analysis. Let $r\leq\min(n,d)$ denote the rank of $X$ and $k\in\mathbb N^+$ such that $1\leq k\leq r$. We  write the SVD of $X\in\R^{d\times n}$ as $X = U_1\Sigma_1V_1^\intercal + U_2\Sigma_2V_2^\intercal$, where $\Sigma_1 \in \R^{k\times k}$, $\Sigma_2 \in \R^{(r-k)\times(r-k)}$, $U_1\in\R^{d\times k}$, $U_2\in\R^{d\times (r-k)}$, $V_1 \in \R^{n\times k}$ and $V_2 \in \R^{n\times (r-k)}$. We use $\sigma_1,\sigma_2,\ldots, \sigma_r$ to denote the singular values of $X$ in the descending order. Let $\mu_k$ denote the coherence measure of $V_1$ defined as $\mu_k = \frac{n}{k}\max_{1\leq i\leq n}\sum_{j=1}^k[V_1]^2_{ij}$.
\begin{thm}[\textbf{RS}~\cite{journals/corr/abs-1110-5305}]
Let $\Omega \in \R^{n \times m}$ be a random sampling matrix corresponding to sampling the columns of $X$ uniformly at random with or without replacement. If for any $\epsilon\in(0,1)$ and $\delta>0$, $m$ satisfies $m \geq \frac{2 \mu_k}{(1-\epsilon)^2}k \log \frac{k}{\delta}$,
then with a probability at least $1-\delta$,
\begin{align*}
\|X -  P_YX\|_2
\leq \sqrt{1 + \frac{n}{\epsilon m}} \sigma_{k+1}
\end{align*}
\end{thm}
{\bf Remark:} The matrix approximation error using RS  implies the excess risk bound  of RS for risk minimization is dominated by  $O(\frac{\sigma_{k+1}}{\sqrt{m}})$ provided $m\geq \Omega(\mu_k k\log k)$, which is in the same order in terms of $m$ to that in Proposition~\ref{prop:1} of oblivious randomized reduction. However, random sampling is not guaranteed to work in oblivious randomized reduction since it does not satisfy the JL lemma in general~\cite{yang2015theory} as required in Proposition~\ref{prop:1}. Moreover, if the data matrix is low rank such that $m\geq \Omega(\mu_r r\log r)$, then the excess risk bound of NOR with RS is $O(1/\sqrt{n})$, the same order to that of $\w_n$ learned from  the original high-dimensional features.

\begin{thm}[\textbf{RG}~\cite{gittens2013revisiting}]
 Let $\Omega \in \R^{n \times m}$ be a random Gaussian matrix. If  for any $\epsilon\in(0,1)$ and $k>4$, $m$ satisfies
$m \geq 2\epsilon^{-2}k\log k$, then with a probability at least $1 - 2k^{-1} - 4k^{-k/\epsilon^2}$,
\begin{align*}
\|X- P_YX\|_2 \leq &O(\sigma_{k+1})+ O\left(\frac{\epsilon}{\sqrt{k\log k}}\right)\sqrt{\sum\nolimits_{j>k}\sigma^2_j}
\end{align*}
\end{thm}
{\bf Remark:} We are interested in comparing the error bound of RG with that of RS. In the worse case, when the tail singular values are flat, then $\|X - P_YX\|_2\leq O(\sqrt{\frac{n}{m}}\sigma_{k+1})$, which is in the same order to that of RS. However, if the tail eigen-values decay fast such that $\sqrt{\sum_{j>k}\sigma^2_j}\ll\sqrt{n}\sigma_{k+1}$, then the matrix approximation error could be much better than $O(\sqrt{\frac{n}{m}}\sigma_{k+1})$, and consequentially the excess risk bound could be much better than $O(\sigma_{k+1}/\sqrt{m})$ that is suffered by RS.

\begin{thm}[\textbf{SRHT}~\cite{DBLP:journals/siammax/BoutsidisG13}]\label{thm:5}
Let $\Omega=\sqrt{\frac{n}{m}}DHP\in\R^{n\times m}$ be a SRHT with $P\in\R^{n\times m}$ being a random sampling matrix, $D\in\R^{n\times n}$ is a diagonal matrix with each entry sampled from $\{1,-1\}$ with equal probabilities and $H\in\R^{n\times n}$ is a normalized Hadamard transform. If for any $0 <\epsilon<1/3$, $2\leq k\leq r$ and $\delta\in(0,1)$,  $m$ satisfies
\begin{align*}
6C^2\epsilon^{-1}[\sqrt{k} + \sqrt{8\log(n/\delta)}]^2\log(k/\delta)\leq m\leq n,
\end{align*}
then with a probability at least $1-5\delta$,
\begin{align*}
\|X- P_YX\|_2&\leq  \left(4+\sqrt{\frac{3\log(n/\delta)\log(r/\delta)}{m}}\right)\sigma_{k+1}\\
&+\sqrt{\frac{3\log(r/\delta)}{m}}\sqrt{\sum\nolimits_{j>k}\sigma_j^2}
\end{align*}
where $C$ is a universal constant.
\end{thm}
{\bf Remark:} The order of the matrix approximation error of SRHT is similar to that of RG up to a logarithmic factor.

Finally, we summarize the matrix approximation error of the RH matrix $\Omega$. This has been studied in~\cite{DBLP:journals/corr/CohenNW15}, in which RH is also referred to as sparse subspace embedding. We first describe the construction  of random hashing matrix $\Omega$. Let $h_k(i):[n]\rightarrow [m/s], k=1,\ldots, s$ denote $s$ independent random hashing functions and let  $\Omega =((H_1D_1)^{\intercal}, (H_2D_2)^{\intercal},\ldots, (H_sD_s)^{\intercal})^{\intercal}\in\R^{m\times n}$ be a random matrix with a block of $s$ random hashing matrices, where $D_k\in\R^{n\times n}$ is a diagonal matrix with each entry sampled from $\{-1, +1\}$ with equal probabilities, and $H_k\in\R^{m/s, n}$ with $[H_k]_{j,i}=\delta_{j, h_k(i)}$. The following theorem below summarizes the matrix approximation error using such a random matrix $\Omega$.
\begin{thm}[\textbf{RH}~\cite{DBLP:journals/corr/CohenNW15}]\label{thm:mat:RH}
 For any $\delta\in(0,1)$ and $\epsilon\in(0,1)$. If $s=1$ and $m=O(k/(\epsilon\delta))$ or $s = O(\log^3(k/\delta)/\sqrt{\epsilon})$ and $m=O(k\log^6(k/\delta)/\epsilon)$, then with a probability $1-\delta$
 \begin{align*}
\|X- P_YX\|_2&\leq  (1+\sqrt{\epsilon})\sigma_{k+1} + \sqrt{\frac{\epsilon}{k}}\sqrt{\sum\nolimits_{j>k}\sigma_j^2}
\end{align*}
\end{thm}
{\bf Remark:} With the second choice of $s$ and $m$,  the order of the matrix approximation error of RH  is similar to that of SRHT up to a logarithmic factor.

To conclude this section, we can see that the excess risk bound of the ERM with non-oblivious randomized reduction is dominated by $O(1/\sqrt{m})$ in the worst case, and could be much better than RG, SRHT and RH if the tail singular values decay fast.

\section{Experiments} \label{experiments}
In this section, we provide empirical evaluations in support of the proposed algorithms and the theoretical analysis. We implement and compare the following algorithms: (i) NOR: ERM with non-oblivious randomized reduction; (ii) previous ERM approaches with oblivious randomized reduction, including two  dual recovery approaches, namely random projection with dual recovery (RPDR)~\cite{DBLP:journals/tit/0005MJYZ14}, and dual-sparse regularized randomized (DSRR) approach~\cite{yang2015theory}, and the pure random projection (RP)~\cite{paul2013random}.
We also implement and compare three randomized reduction operators for these different approaches, i.e., RH, RG and RS~\footnote{We do not report the performance of SRHT because it has similar performance to RH but it is less efficient than RH. }. For RH, we use only one block of random hashing matrix (i.e., $s=1$). A similar result to Therorem~\ref{thm:mat:RH} can be established for one-block of random hashing but with a constant success probability~\cite{DBLP:journals/corr/abs-1211-1002}. The loss function for the binary classification problem is the hinge loss and for the multi-class classification problem is the softmax loss.

\begin{table}[t]
\caption{Statistics of real datasets}
\label{table1}
\begin{center}
\begin{small}
\begin{tabular}{lccccr}
\hline
Name & \#Training & \#Testing & \#Features & \#Classes \\
\hline
RCV1.b  & 677,399 & 20,242 & 47,236 &2 \\
Splice & 1,000,000 & 4,627,840 & 12,495,340 &2 \\
RCV1.m & 15,564 & 518,571 & 47,236 &53 \\
News20 & 15,935 & 3,993 & 62,061 &20 \\
\hline
\end{tabular}
\end{small}
\end{center}
\end{table}
\begin{figure}[t]
\begin{center}
\subfigure[RPDR and DSRR]{\hspace*{-0.05in}\includegraphics[scale=0.45]{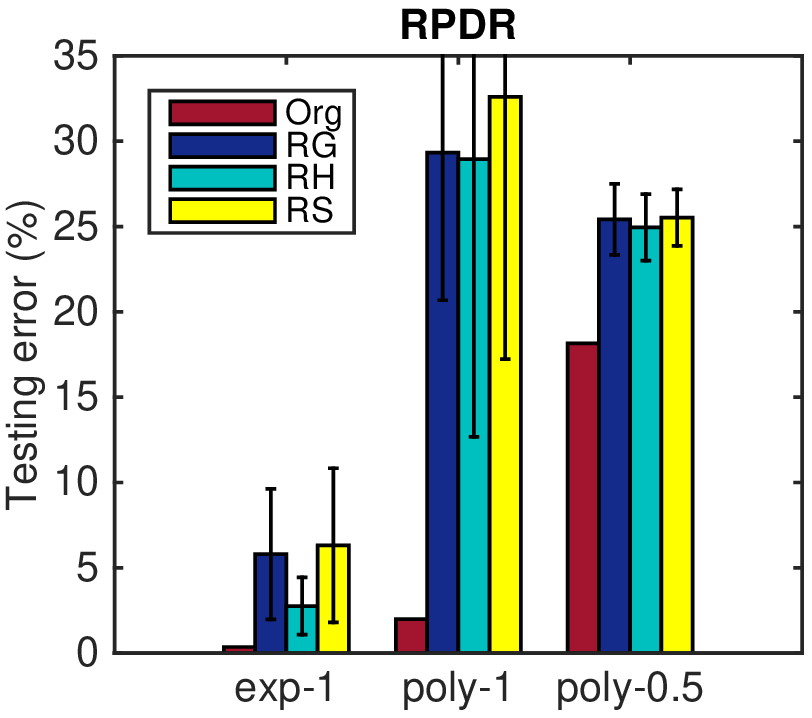}\hspace*{-0.07in}\includegraphics[scale=0.45]{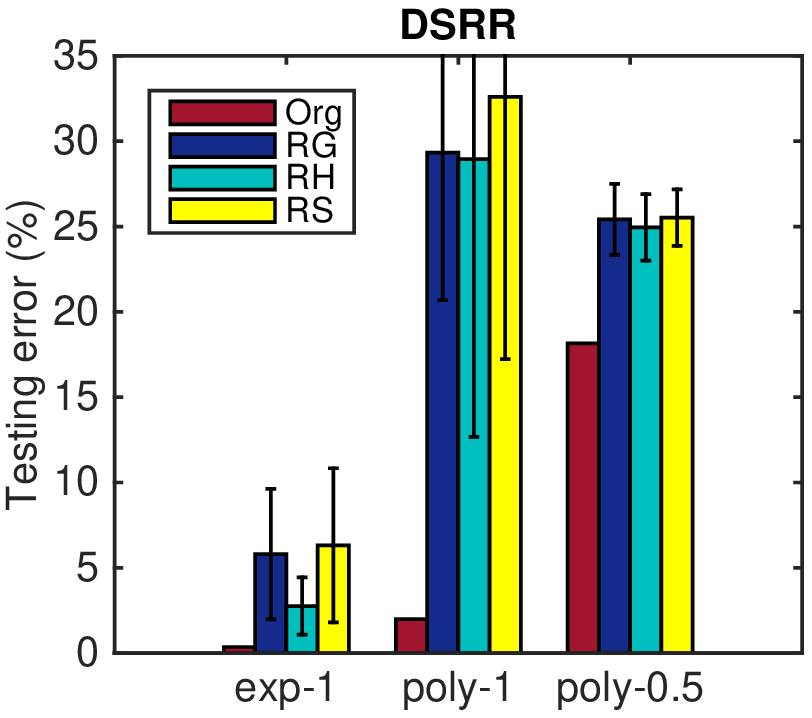}}\hspace*{-0.1in}\\
\subfigure[NOR]{\includegraphics[scale=0.45]{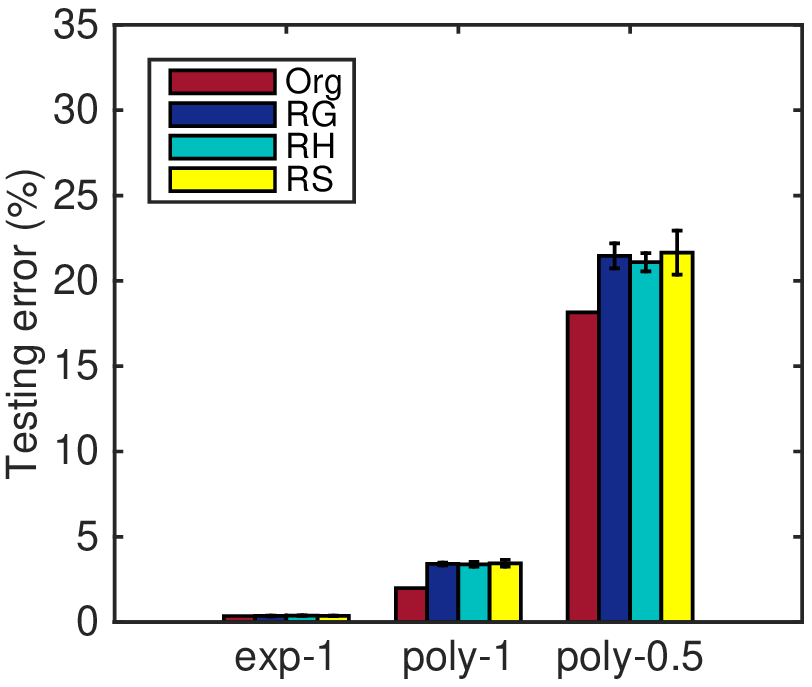}}\hspace*{-0.1in}
\subfigure[RH]{\includegraphics[scale=0.45]{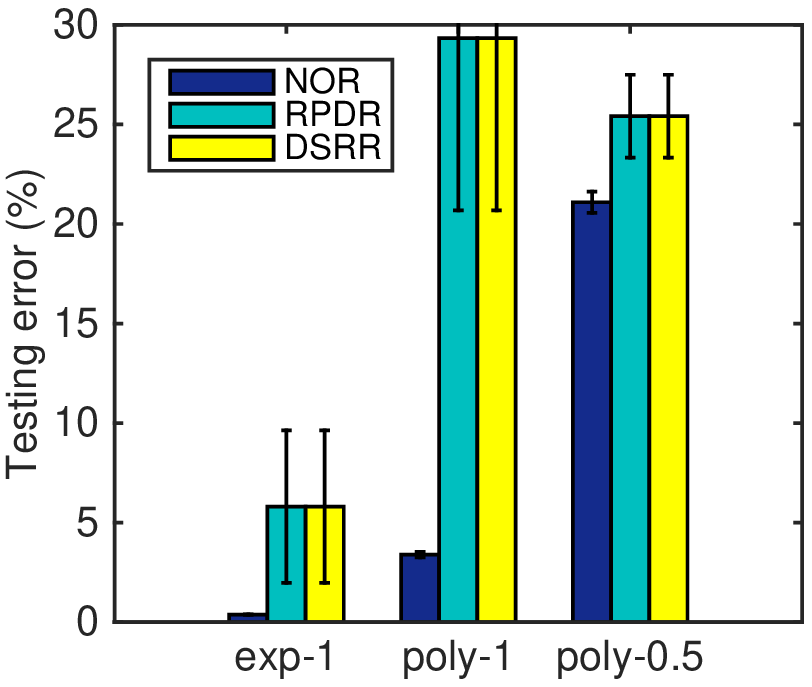}}
\caption{ (a) RPDR and DSRR with different randomized reduction operators. (b) NOR with different randomized reduction operators. (c) Different approaches with RH. }
\label{fig1}
\end{center}
\end{figure}
\begin{figure}[t]
\begin{center}
\includegraphics[scale=0.345]{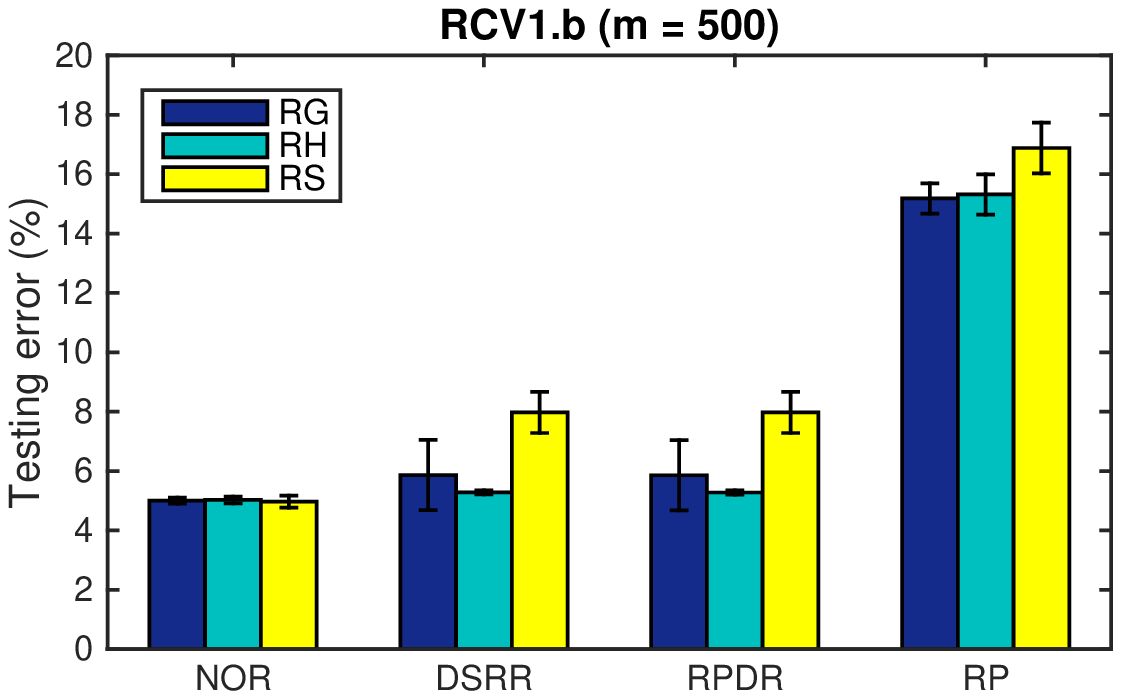}\hspace*{-0.15in}
\includegraphics[scale=0.345]{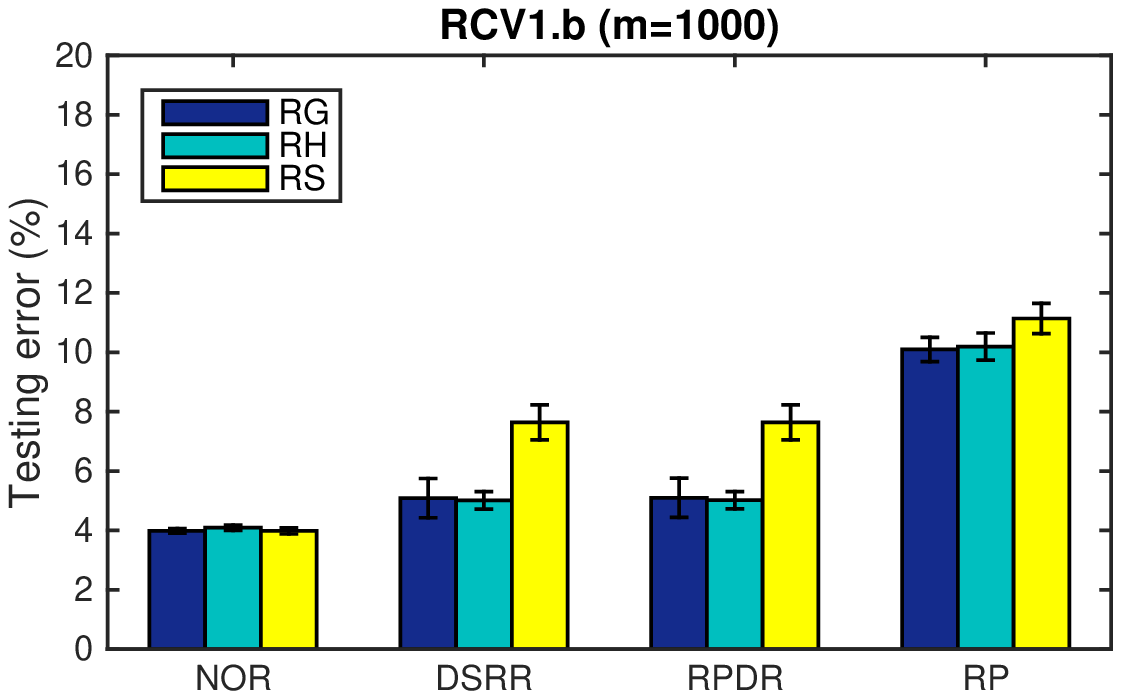}\hspace*{-0.15in}\\
\includegraphics[scale=0.345]{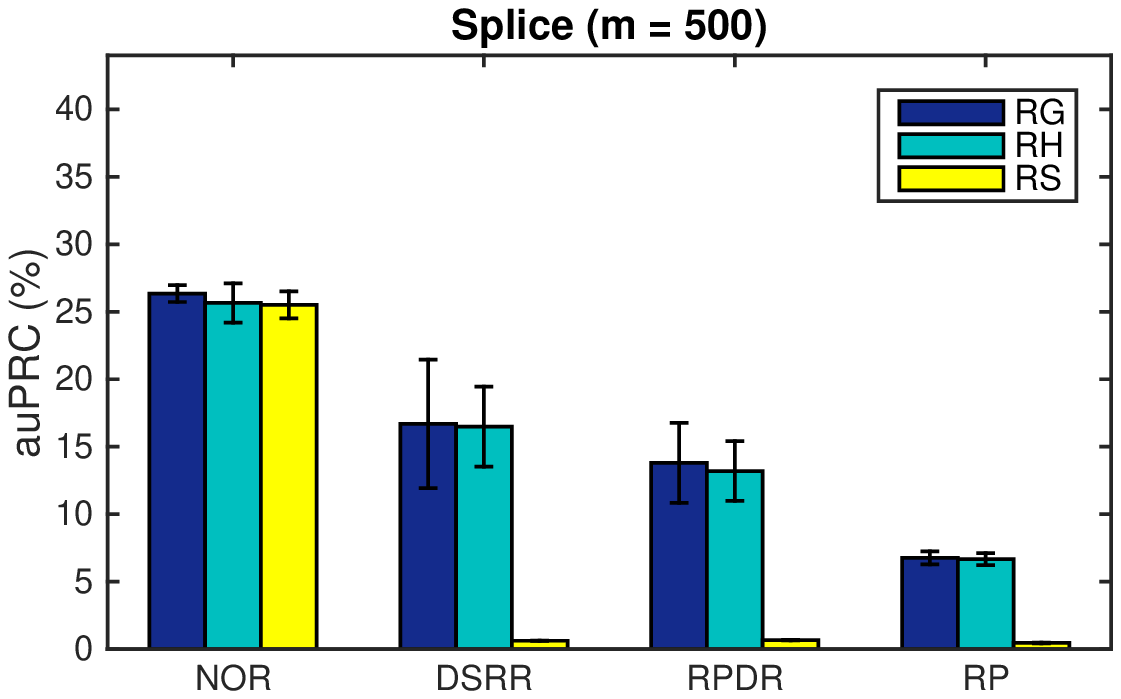}\hspace*{-0.15in}
\includegraphics[scale=0.345]{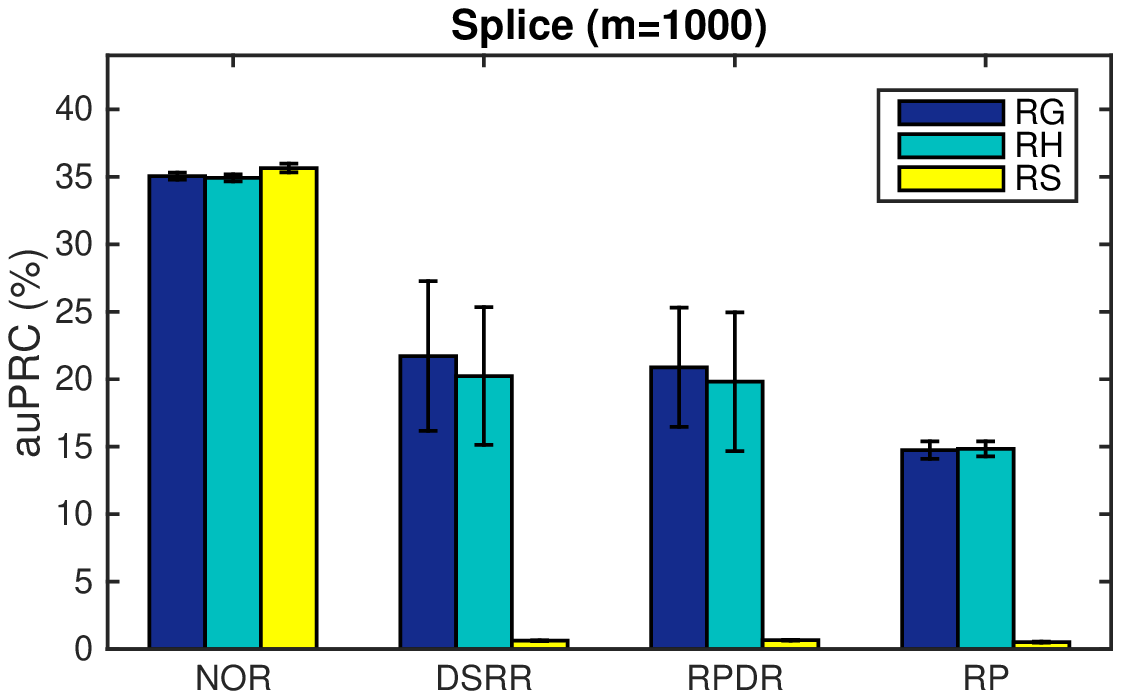}
\caption{Testing performance of different approaches on RCV1.b and Splice datasets.}
\label{fig2}
\end{center}
\end{figure}
\begin{figure}[t]
\begin{center}
\includegraphics[scale=0.3]{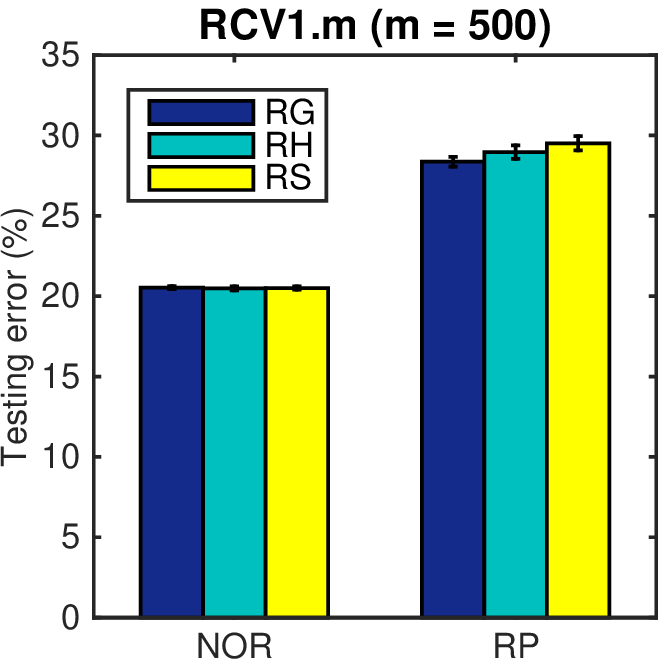}\hspace*{-0.09in}
\includegraphics[scale=0.3]{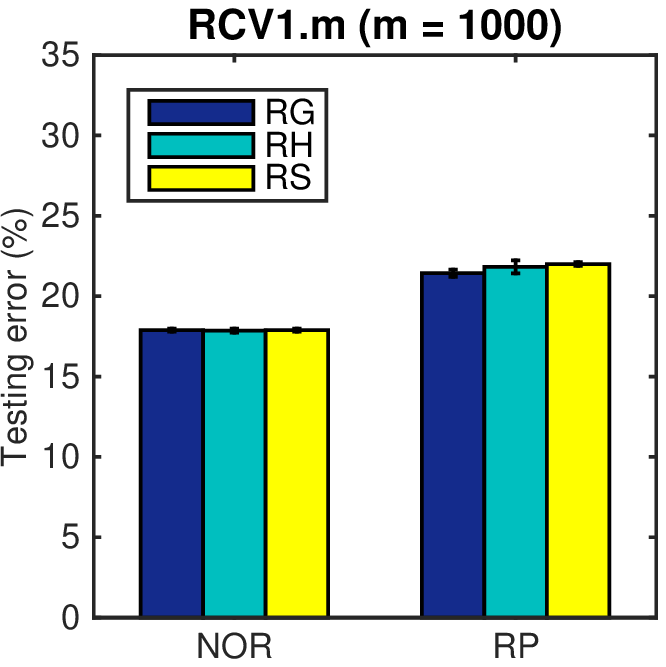}\hspace*{-0.09in}
\includegraphics[scale=0.3]{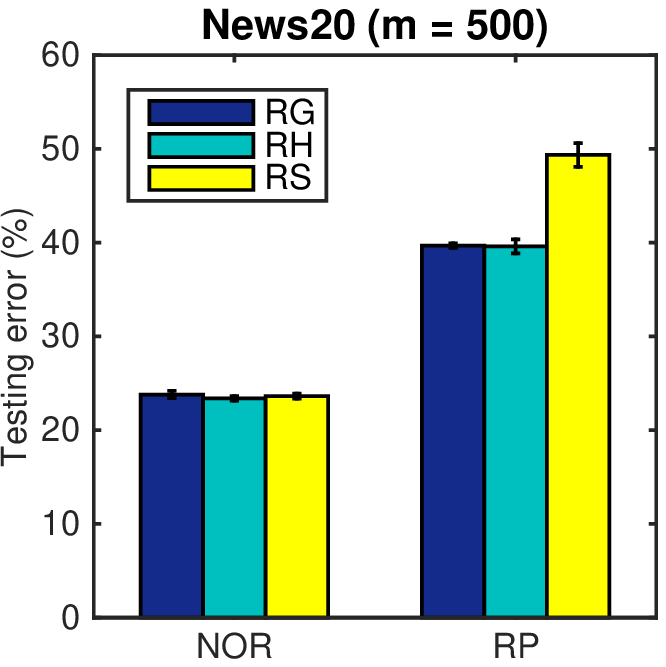}\hspace*{-0.09in}
\includegraphics[scale=0.3]{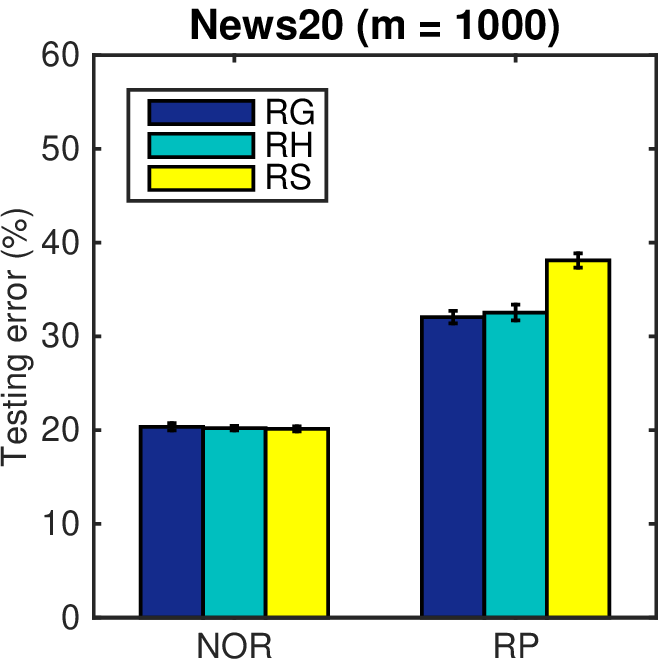}
\includegraphics[scale=0.3]{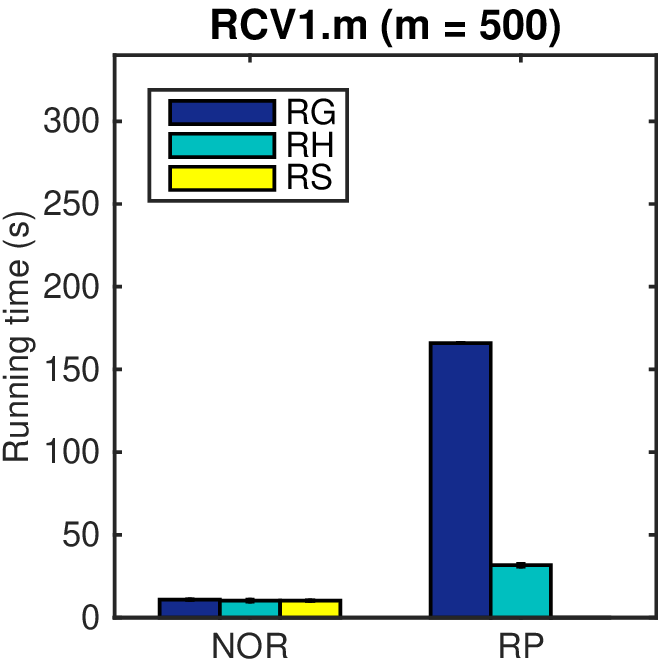}\hspace*{-0.09in}
\includegraphics[scale=0.3]{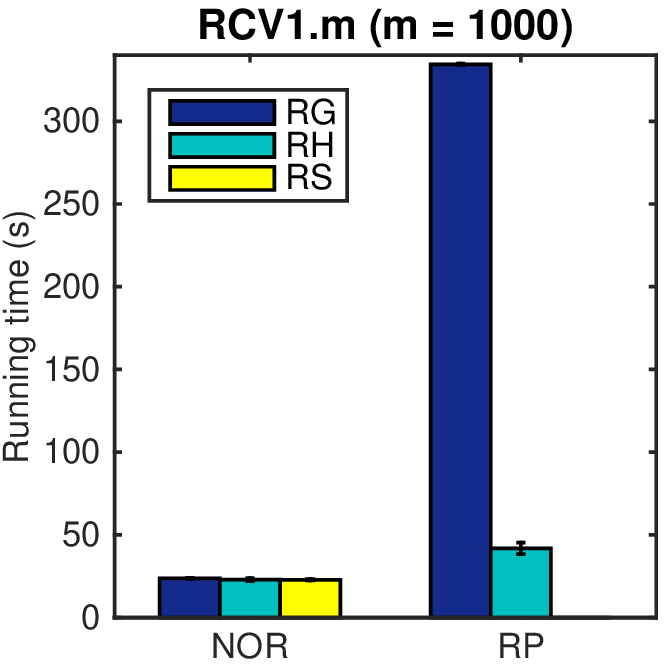}\hspace*{-0.09in}
\includegraphics[scale=0.3]{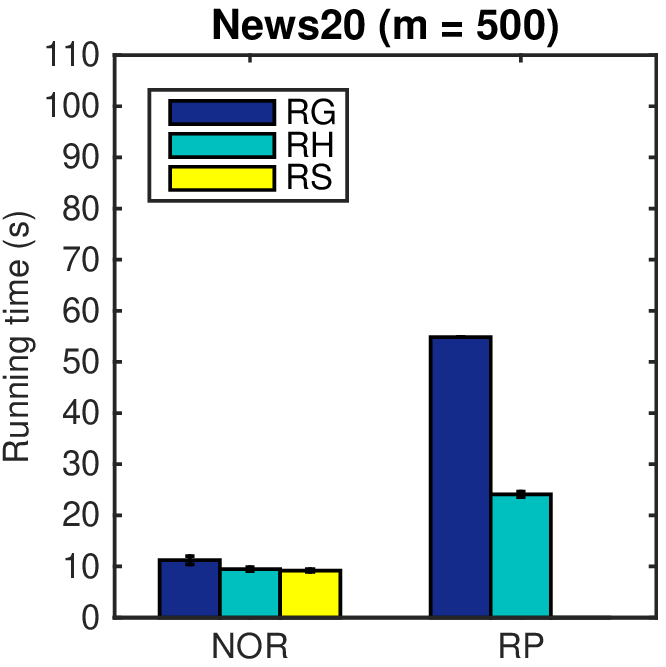}\hspace*{-0.09in}
\includegraphics[scale=0.3]{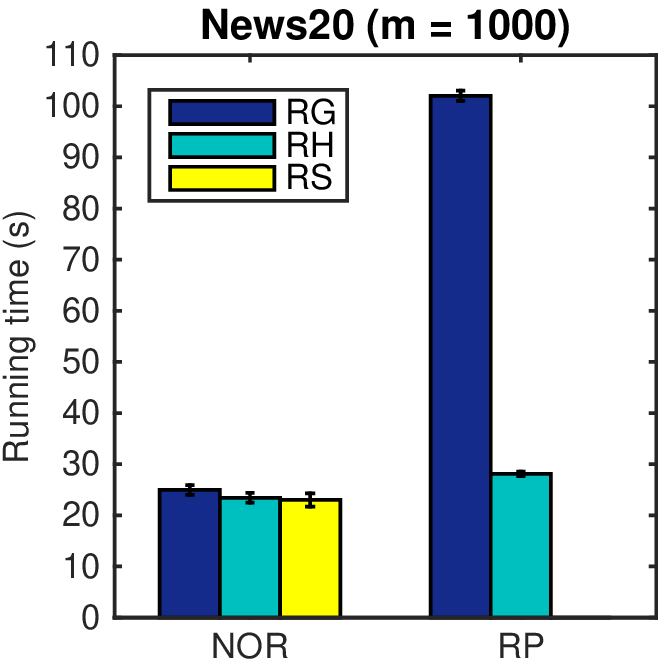}
\caption{Testing performance  and running time of different approaches on RCV1.m and News20 datasets.}
\label{fig3}
\end{center}
\end{figure}

Experiments are conducted  on  four real-world datasets and three synthetic datasets. The four real-world datasets are described in Table~\ref{table1}. To generate synthetic data, we first draw a random standard Gaussian matrix $M \in \R^{d \times n}$ and compute its SVD $M = USV^\intercal$. Then we construct singular values following three different decay:  an exponential decay (exp-$\tau$) with  $\sigma_{i} = e^{-i\tau}, (\tau = 1)$ and polynomial decay (poly-$\tau$) with  $\sigma_{i} = i^{-\tau}, (\tau = 0.5, 1)$. This will generate 3 synthetic datasets. We compute a base data matrix by $X_b = \sqrt{n}U \Sigma V^\intercal$, where $\Sigma = \diag \{\sigma_1, \dots, \sigma_d\}$. Then the binary labels are computed by $\y = sign(X_b^\intercal \w)$, where $\w \in \R^{d}$ is a standard Gaussian random vector. To increase the difficulty of the problem, we add some Gaussian random features to each data in $X_b$ and form a full data matrix $X \in \R^{(d+t) \times n}$. We use the first $90\%$ examples as training data and the remaining  $10\%$ examples as testing data. In particular, we generate the synthetic datasets with $d = 1000, n = 10^5, t =10$. We note that the synthetic data is not high-dimensional and it is solely for verifying the proposed approach and analysis. We perform data reduction to reduce features to the dimensionality of $m = 100$.

We first compare the performance of NOR, RPDR  and DSRR with different randomized reduction operators on the synthetic datasets in order to verify the excess risk bounds established  in Proposition~\ref{prop:1} and subsection ``Matrix Approximation Error" (MAE).   The results are shown in Figure~\ref{fig1}, where we also include the performance of SVM on the original features (denoted by Org). From the results, we can observe that  (i) when the singular values follow an exponential decay (which yields almost low-rank data matrices), NOR performs almost the same to SVM on the original data; however the two recovery approaches RPDR and DSRR perform much worse than SVM, verifying our theoretical analysis in Proposition~\ref{prop:1} and subsection ``MAE"; (ii) The performance of NOR decreases gradually as the decay of singular values becomes slower, which is consistent with the theoretical results in subsection ``MAE"; (iii) for NOR,  RS is comparable to RH and RG when the decay of singular-values is fast, but is slightly worse when the decay of singular values becomes slower. This is also expected according to the discussions in subsection ``MAE"; (iv) NOR always performs better than RPDR and DSRR. One reason that RPDR and DSRR do not perform well on these synthetic data sets is that the recovered dual solution is not  accurate because that the data has noise. In contrast, NOR is much more robust to noise.

Secondly, we present some experimental results on two binary classification real datasets, namely Reuters Text Categorization both for binary version (RCV1.b) and Splice Site Recognition (Splice), which are tested in previous studies~\cite{sonnenburg2010coffin}.  For Splice dataset, we evaluate different algorithms by computing the same measure, namely area under precision recall curve (auPRC), as in~\cite{sonnenburg2010coffin}. We compare NOR with RPDR, DSRR and RP. The results are shown in Figure~\ref{fig2}. We can see that when $m$ increases, the testing error/auPRC is monotonically decreasing/increasing. Comparing with other three algorithms, NOR has the best performance. In addition, RS does not work well for oblivious randomized reduction approaches (RP, RPDR and DSRR), but performs similarly to other randomized operators for NOR, which is consistent with our analysis (i.e., RS is not a JL transform as required in Proposition~\ref{prop:1} for RPDR and DSRR; however, RS provides guarantee on the matrix approximation error that renders NOR work).

Finally, we compare NOR with RP on RCV.m and News20 datasets for multi-class classification. The results are shown in Figure~\ref{fig3} (upper panel). We can see that NOR clearly outperforms RP. In addition, running time results \footnote{We do not report the running time for RP using RS as randomized reduction operator since it has the worst performance} are reported in Figure~\ref{fig3} (lower panel). The running time consists of the reducation time and the optimization time in the reduced feture space.
The results show that (i) NOR is more efficient than RP; (ii) RH and/or RS are much more efficient than RG for a certain approach.
It is interesting to note that the total running time of NOR is less than RP. The reason is that the optimization of NOR is more efficient than RP~\footnote{We terminate the optimization for both methods by the same criterion, i.e., the duality gap is less than $10^{-3}$.} due to that the new data of NOR is better suited for classification (higher prediction performance), making the optimization easier, though NOR has slightly higher data reduction time than RP.

\section{Conclusions}
In this paper, we have established the excess risk bound of non-oblivious randomized reduction method for risk minimization problems. More importantly, the new excess risk bound does not require stringent  assumptions of the data and the loss functions, which is nontrivial and significant theoretical results. The empirical studies on synthetic datasets and real datasets validate our theoretical analysis and also demonstrate the effectiveness of the proposed non-oblivious randomized reduction approach.

\section{Acknowlegements}
We thank the anonymous reviewers for their helpful comments. Y. Xu and T. Yang are partially supported by National Science Foundation (IIS-1463988, IIS-1545995).
L. Zhang is partially supported by NSFC (61603177) and JiangsuSF (BK20160658).

\bibliographystyle{aaai}
\bibliography{reference}

\section*{Appendix}
\subsection{Proof of Lemma 2}
Since the loss function is $G$-Lipschitz continuous, i.e., $|\ell'(z,y)|\leq G$, therefore, $\max_{\alpha\in\Omega}|\alpha|\leq G$.
Our proof is built on the dual formulation. First, we have
\begin{align*}
F(\w_n) &= \max_{\alpha\in\Omega^n}- \frac{1}{n}\sum_{i=1}^n\ell^*_i(\alpha_i) - \frac{1}{2\lambda n^2}\alpha^{\top}X^{\top}X\alpha\\
F(\wh_n) &= \max_{\alpha\in\Omega^n}- \frac{1}{n}\sum_{i=1}^n\ell^*_i(\alpha_i) - \frac{1}{2\lambda n^2}\alpha^{\top}X^{\top}\widehat U\widehat U^{\top}X\alpha
\end{align*}
Then
\begin{align*}
F(\wh_n) &= \max_{\alpha\in\Omega^n}- \frac{1}{n}\sum_{i=1}^n\ell^*_i(\alpha_i) - \frac{1}{2\lambda n^2}\alpha^{\top}X^{\top}\widehat U\widehat U^{\top}X \alpha\\
&= \max_{\alpha\in\Omega^n}- \frac{1}{n}\sum_{i=1}^n\ell^*_i(\alpha_i) - \frac{1}{2\lambda n^2}\alpha^{\top}X^{\top}X\alpha \\
&+ \frac{1}{2\lambda n^2}\alpha^{\top}(X^{\top}X - X^{\top}\widehat U\widehat U^{\top}X)\alpha\\
&\leq \max_{\alpha\in\Omega^n}- \frac{1}{n}\sum_{i=1}^n\ell^*_i(\alpha_i) - \frac{1}{2\lambda n^2}\alpha^{\top}X^{\top}X\alpha \\
& + \max_{\alpha\in\Omega^n}\frac{1}{2\lambda n^2}\alpha^{\top}(X^{\top}X -  X^{\top}\widehat U\widehat U^{\top}X)\alpha
\end{align*}
\begin{align*}
&\leq F(\w_n) \\
& + \frac{1}{2\lambda n^2}\max_{\alpha\in\Omega^n}\|\alpha\|_2^2\|X^{\top}X -  X^{\top}\widehat U\widehat U^{\top}X\|_2\\
&\leq F(\w_n) + \frac{G^2}{2\lambda n}\|X^{\top}X -  X^{\top}\widehat U\widehat U^{\top}X\|_2
\end{align*}
On the other hand, since $ P_Y  = \widehat U\widehat U^{\top}$ is the projection to the column space of $Y=X\Omega$, we have
\begin{align*}
&\|X^{\top}X -  X^{\top}\widehat U\widehat U^{\top}X\|_2 = \|X^{\top}X - X^{\top}P_{Y}X\|_2 \\
&= \|X^{\top}(I - P_{Y})X\|_2 = \|X^{\top}(I - P_{Y})^2X\|_2  \\
& \ \ \text{(by the property of projection)} \\
&= \|X - P_{Y}X\|_2^2
\end{align*}
Combing above results together, we complete the proof.

\subsection{Proof of Theorem 1}
\begin{proof}

\begin{align*}
\text{ER}&(\wh_n, \w_*)  = \E_{\P}[\ell(\wh_n^{\top}\x, y)] - \E_{\P}[\ell(\w_*^{\top}\x, y)] \\
& =\left[ \bar F(\wh_n) - \frac{\lambda}{2}\|\wh_n\|_2^2 \right]- \left[ \bar F(\w_*) - \frac{\lambda}{2}\|\w_*\|_2^2 \right] \\
& = \bar F(\wh_n) - \bar F(\w_*) + \frac{\lambda}{2} \|\w_*\|_2^2 - \frac{\lambda}{2}\|\wh_n\|_2^2 \\
& \leq \bar F(\wh_n) - \bar F(\w_*) + \frac{\lambda}{2} \|\w_*\|_2^2 \\
& \leq (1+a) (F(\wh_n) - F(\w_n)) \\
& + \frac{8(1+1/a)G^2R^2(32 + \log(1/\delta))}{\lambda n} + \frac{\lambda}{2} \|\w_*\|_2^2\\
&  \ \ \text{(by Lemma 3)}\\
& \leq \frac{G^2(1+a)}{2\lambda n}\|X- P_YX\|^2_2 \\
& + \frac{8(1+1/a)G^2R^2(32 + \log(1/\delta))}{\lambda n} + \frac{\lambda}{2} \|\w_*\|_2^2 \\
& \ \text{(by Lemma 2)} \\
& \leq \frac{G^2(1+a)}{2\lambda n}\|X- P_YX\|^2_2 \\
& + \frac{8(1+1/a)G^2R^2(32 + \log(1/\delta))}{\lambda n} + \frac{\lambda B^2}{2}
\end{align*}
The last inequality is held because of $\|\w_*\|_2^2 \leq B^2$.
\end{proof}

\subsection{Proof of Theorem 2}
\begin{proof}
Recall that $X = U_1\Sigma_1V_1^\top + U_2\Sigma_2V_2^\top$, and we have  $Y = X\Omega = U_1\Sigma_1V_1^\top\Omega + U_2\Sigma_2V_2^\top \Omega$. Let's denote $\Omega_1 = V_1^\top\Omega$ and  $\Omega_2 = V_2^\top\Omega$, then
\begin{align*}
Y = X\Omega = U_1\Sigma_1\Omega_1 + U_2\Sigma_2\Omega_2
\end{align*}
Based on the results from Theorem 9.1 of ~\cite{halko2011finding}, we have
\begin{align}\label{SpecBound}
\|X- P_Y X\|_2^2  \leq \| \Sigma_2\|_2^2 + \| \Sigma_2 \Omega_2 \Omega_1^{\dag}\|_2^2
\end{align}
with a probability at least $1-\delta$.
In~\cite{journals/corr/abs-1110-5305}, Lemma 1 showed that
\begin{align*}
 \| \Omega_1^{\dag}\|_2^2 \leq \frac{n}{\varepsilon m}
\end{align*}
if assume that $\Omega_1$ has full row rank. It is easy to show that
$\| \Omega_2 \|_2^2 \leq \|V^{\top}_2 \|_2^2 \| \Omega \|_2^2 \leq 1$. Then we have
\begin{align*}
\| \Sigma_2 \Omega_2 \Omega_1^{\dag}\|_2^2 \leq \frac{n}{\varepsilon m} \| \Sigma_2 \|_2^2
\end{align*}
Combining this result with equation (\ref{SpecBound}), we know that
\begin{align*}
\|X- P_Y X\|_2^2  \leq (1+\frac{n}{\varepsilon m}) \| \Sigma_2\|_2^2
\end{align*}
i.e.
\begin{align*}
\|X- P_Y X\|_2  \leq \sqrt{(1+\frac{n}{\varepsilon m})} \sigma_{k+1}
\end{align*}
with a probability at least $1-\delta$.
\end{proof}

\subsection{Proof of Theorem 3}
\begin{proof}
Our proof is modified from~\cite{gittens2013revisiting}. In Section 10 of~\cite{halko2011finding}, it is showed that
if $m = k + p$ with $p > 4$ and $u, t \geq 1$, and $\Sigma_2$ is a diagonal matrix, then we have
\begin{align}\label{Prob}
\nonumber \| \Sigma_2 \Omega_2 \Omega_1^{\dag}\|_2 &\leq \| \Sigma_2 \|_2 \left( \sqrt{\frac{3k}{p+1}} t+ \frac{e\sqrt{m}}{p+1} tu \right) \\
&+ \| \Sigma_2 \|_F \frac{e\sqrt{m}}{p+1} t
\end{align}
with probability at least $1-2t^{-p} - e^{-u^2/s}$.
Since $m \geq 2\epsilon^{-2}k\log k$ and $m = k + p$, we have that $p \geq \epsilon^{-2}k\log k$. Then, the following inequalities hold:
\begin{align*}
 &\sqrt{\frac{3k}{p+1}} \leq \sqrt{\frac{3k}{p}} \leq \sqrt{\frac{3}{\log k}} \epsilon \\
 &\frac{\sqrt{m}}{p+1} \leq  \frac{\sqrt{k+p}}{p} \leq \sqrt{\frac{\epsilon^{4}}{k\log^2 k} + \frac{\epsilon^{2}}{k\log k} } \leq \sqrt{\frac{2}{k\log k}} \epsilon \\
\end{align*}
Apply these inequalities and set $t=e$ and $u=\sqrt{2 \log k}$ in (\ref{Prob}), we can obtain

\begin{align*}
 \| \Sigma_2 \Omega_2 \Omega_1^{\dag}\|_2 & \leq \| \Sigma_2 \|_2 \left( e\sqrt{\frac{3k}{p+1}} + \frac{e^2\sqrt{2m \log k} }{p+1} \right) \\
 &+ \| \Sigma_2 \|_F \frac{e^2\sqrt{m}}{p+1} \\
 & \leq \left( e\sqrt{\frac{3}{\log k}}  + 2e^2 \sqrt{\frac{1}{k}} \right)  \epsilon \| \Sigma_2 \|_2 \\
 &+  e^2\sqrt{\frac{2}{k\log k}} \epsilon \| \Sigma_2 \|_F
\end{align*}
Combining this results into equation (\ref{SpecBound}), we have
\begin{align*}
\|X- P_Y X\|_2^2  & \leq \| \Sigma_2\|_2^2 + \| \Sigma_2 \Omega_2 \Omega_1^{\dag}\|_2^2 \\
& \leq \| \Sigma_2\|_2^2 + \left( e\sqrt{\frac{3}{\log k}}  + 2e^2 \sqrt{\frac{1}{k}} \right)^2  \epsilon^2 \| \Sigma_2 \|_2^2 \\
&+  e^4\frac{2}{k\log k} \epsilon^2 \| \Sigma_2 \|_F^2 \\
& \leq \left[ 1+ \left( e\sqrt{\frac{3}{\log k}}  + 2e^2 \sqrt{\frac{1}{k}} \right)^2  \epsilon^2 \right] \sigma_{k+1}^2 \\
&+  e^4\frac{2}{k\log k} \epsilon^2 {\sum_{j>k}\sigma^2_j}
\end{align*}
Thus
\begin{align*}
\|X- P_Y X\|_2  &  \leq \sqrt{ 1+ \left( e\sqrt{\frac{3}{\log k}}  + 2e^2 \sqrt{\frac{1}{k}} \right)^2  \epsilon^2 } \sigma_{k+1} \\
&+  e^2\sqrt{\frac{2}{k\log k}} \epsilon \sqrt{\sum_{j>k}\sigma^2_j} \\
& \leq O(\sigma_{k+1})+ O\left(\frac{\epsilon}{\sqrt{k\log k}}\right)\sqrt{\sum_{j>k}\sigma^2_j}
\end{align*}
with a probabilty at least $1 - 2k^{-1} - 4k^{-k/\epsilon^2}$.
\end{proof}

\subsection{Proof of Theorem 4}
\begin{proof}
In~\cite{boutsidis2013improved}, Lemma 4.1 showed that
\begin{align*}
 \| \Omega_1^{\dag}\|_2^2 \leq (1-\sqrt{\epsilon})^{-1}
\end{align*}
with probability at least $1-3\delta$. Comsequently, $\Omega_1$ has full row rank, and by applying Lemma 5.4 of ~\cite{boutsidis2013improved} with the same probability, we obtain
\begin{align}\label{Th6}
\|X- P_Y X\|_2^2  \leq  \| \Sigma_2\|_2^2 + (1-\sqrt{\epsilon})^{-1} \| \Sigma_2 V_2^{\top} \Omega\|_2^2
\end{align}
From Lemma 4.8 of ~\cite{boutsidis2013improved} we have
\begin{align*}
&(1-\sqrt{\epsilon})^{-1} \| \Sigma_2 V_2^{\top} \Omega\|_2^2  \leq \frac{5}{1-\sqrt{\epsilon}} \| \Sigma_2 V_2^{\top} \|_2^2 \\
&+ \frac{\log (r/\delta)}{(1-\sqrt{\epsilon})m} (\| \Sigma_2 V_2^{\top} \|_F + \sqrt{8 \log(n/\delta)}\| \Sigma_2 V_2^{\top} \|_2)^2
\end{align*}
with probability at least $1-5\delta$. Since $0 < \epsilon < 1/3$, then $(1-\sqrt{\epsilon})^{-1} < 3$. Also
$\| \Sigma_2 V_2^{\top} \|_2 = \| \Sigma_2 \|_2$ and $\| \Sigma_2 V_2^{\top} \|_F = \| \Sigma_2 \|_F$. Thus,
\begin{align*}
&(1-\sqrt{\epsilon})^{-1} \| \Sigma_2 V_2^{\top} \Omega\|_2^2  \leq 15 \| \Sigma_2 \|_2^2 \\
&+ \frac{3\log (r/\delta)}{m} (\| \Sigma_2 \|_F + \sqrt{8 \log(n/\delta)}\| \Sigma_2 \|_2)^2
\end{align*}
 Plugging this equation into (\ref{Th6}),
 \begin{align*}
\|X- P_Y X\|_2^2  &\leq 16 \| \Sigma_2 \|_2^2 + \\
&\frac{3\log (r/\delta)}{m} (\| \Sigma_2 \|_F + \sqrt{8 \log(n/\delta)}\| \Sigma_2 \|_2)^2
\end{align*}
Use the subadditivity of the square-root function to obtain that
  \begin{align*}
\|X- P_Y X\|_2  
& \leq \left(4 + \sqrt{\frac{3\log(n/\delta)\log (r/\delta)}{m}} \right) \sigma_{k+1} \\
&+ \sqrt{\frac{3\log (r/\delta)}{m}} \sqrt{\sum_{j>k}\sigma^2_j}
\end{align*}
with probability at least $1-5\delta$.
\end{proof}

\subsection{Proof of Theorem 5}
\begin{proof}
Recall that $X=U_1 \Sigma_1 V_1^\top + U_2 \Sigma_2 V_2^\top$ and $Y=X\Omega=\widehat U\widehat\Sigma \widehat V^{\top}$.
Applying the SVD of $P_Y X=\widehat U_k\widehat\Sigma_k \widehat V_k^{\top} + \widehat U_{\bar k} \widehat\Sigma_{\bar k} \widehat V_{\bar k}^{\top}$ to Theorem 4 of~\cite{DBLP:journals/corr/CohenNW15}, we have
\begin{align*}
&\|X- \widehat U_k\widehat\Sigma_k \widehat V_k^{\top}\|_2^2 \\
& \leq (1+\epsilon)\| X - U_1 \Sigma_1 V_1^\top\|_2^2 + \frac{\epsilon}{k} \| X - U_1 \Sigma_1 V_1^\top\|_F^2 \\
& = (1+\epsilon)\| \Sigma_2 \|_2^2 + \frac{\epsilon}{k} \| \Sigma_2\|_F^2
\end{align*}
with probability at least $1-\delta$. Then we have
\begin{align*}
\|X- P_Y X\|_2^2 & = \|X- \widehat U_k\widehat\Sigma_k \widehat V_k^{\top} - \widehat U_{\bar k} \widehat\Sigma_{\bar k} \widehat V_{\bar k}^{\top}\|_2^2 \\
& \leq \|X- \widehat U_k\widehat\Sigma_k \widehat V_k^{\top}\|_2^2 + \| \widehat U_{\bar k} \widehat\Sigma_{\bar k} \widehat V_{\bar k}^{\top}\|_2^2 \\
&\leq \|X- \widehat U_k\widehat\Sigma_k \widehat V_k^{\top}\|_2^2 \\
& \leq (1+\epsilon)\| \Sigma_2 \|_2^2 + \frac{\epsilon}{k} \| \Sigma_2\|_F^2
\end{align*}
We complete the proof by using the subadditivity of the square-root function, i.e. we have
\begin{align*}
\|X- P_Y X\|_2 &\leq \sqrt{(1+\epsilon)}\| \Sigma_2 \|_2 + \sqrt{\frac{\epsilon}{k}} \| \Sigma_2\|_F \\
& \leq (1+\sqrt{\epsilon})\sigma_{k+1} + \sqrt{\frac{\epsilon}{k}}\sqrt{\sum_{j>k}\sigma_j^2}
\end{align*}
with probability at least $1-\delta$.
\end{proof}

\subsection{An Efficient Implementation of Computing $\widehat U$ in NOR for Sparse Data }
In this section, we present an efficient implementation of NOR for sparse data.
We note that when data is sparse, the time complexity of calculating $\widehat U^{\top}X$ is $O(mN)$, where $N \ll dn$ is the number of non-zero elements in $X$. Consequently, computing the left singular vectors of $Y$ could become a significant component of overall computation in Algorithm 2. To harness data sparsity, next we present a fast implementation of $\widehat U$ computation. Let $Y=\widehat U\widehat\Sigma \widehat V^{\top}$ be the SVD of $Y$, then $\widehat V\widehat\Sigma^2 \widehat V^{\top}$ is the
singular value decomposition of $K_m = Y^\top Y \in \R^{m \times m}$. Then, the projection matrix $\widehat U^{\top}$ can be computed by
\begin{align} \label{FasterU}
\widehat U^{\top} = \widehat\Sigma^{-1} \widehat V^\top Y^\top
\end{align}
 Therefore, we can efficiently implement the projection matrix by first computing singular values and corresponding left singular vectors of the small matrix $K_m = Y^\top Y \in \R^{m \times m}$ and then compute the projection matrix by Eq. (\ref{FasterU}). We assume the number of non-zero entries in $X$ is $N$ and the number of non-zero entries in $Y$ is $N_m$. The time complexity consists of (i) $O(mN_m)$ for computing $K_m$, (ii) $O(m^2 \log m)$ for computing left singualr value decomposition of $K_m$ by randomized algorithms~\cite{halko2011finding}, (iii) $O(m^2 + mN_m)$ for computing $\widehat\Sigma^{-1} \widehat V^\top Y^\top$, yeilding an overall time complexity of $O(m^2 \log m +m^2 + 2mN_m)$. Compared to the overall time complexity of $O(md\log m)$ by directly computing the left singular vectors of $Y$, the efficient implementation could be much faster especially when $N_m, m \ll d$. We present the detailed steps in \textbf{Algorithm}~\ref{alg:FPM} for computing $\widehat U^{\top}$.

\begin{algorithm}[tb]
\small
   \caption{Fast Projection in NOR}
   \label{alg:FPM}
\begin{algorithmic}[1]
   \STATE Compute $K_m = Y^\top Y \in \R^{m \times m}$
   \STATE Compute eigen-values $\lambda_i $ and corresponding  eigen-vectors $V_i (i = 1, \dots, m)$ of the small matrix $K_m$
   \STATE Let $\widehat\Sigma = diag(\sqrt{\lambda_1},\ldots, \sqrt{\lambda_m})$
   \STATE Compute the projection matrix $\widehat U^{\top}$ by Eq. (\ref{FasterU})
\end{algorithmic}
\end{algorithm}



\end{document}